\documentclass[10pt,conference]{IEEEtran}
\usepackage{amsmath,amssymb}
\usepackage{booktabs}
\usepackage{graphicx}
\usepackage[hidelinks]{hyperref}
\usepackage{cite}
\usepackage{microtype}
\usepackage{xcolor}
\usepackage{tikz}
\usepackage{pgfplots}
\usepackage[per-mode=symbol]{siunitx}
\usepackage{stfloats}
\usepackage{placeins}
\usetikzlibrary{arrows.meta,positioning,fit,backgrounds,calc,shapes.geometric}
\pgfplotsset{compat=1.18}

\definecolor{nxblue}{HTML}{1F4E79}
\definecolor{nxteal}{HTML}{2A9D8F}
\definecolor{nxamber}{HTML}{E9A319}
\definecolor{nxred}{HTML}{C1432E}
\definecolor{nxgray}{HTML}{5B6670}

\newcommand{\code}[1]{\texttt{\small #1}}

\begin{document}

\title{Nexus: Depth-Adaptive KV-Cache Splicing and\\
Retrieval-Decoupled Tool Routing for Agentic LLMs\\
on Unified Memory}

\author{\IEEEauthorblockN{Mustafa Arslan}
\IEEEauthorblockA{\textit{Independent Researcher} \\
Istanbul, T\"urkiye \\
mustafarslan35@gmail.com}}

\maketitle

\begin{abstract}
Agentic large language models (LLMs) on the Model Context Protocol (MCP) re-encode verbose tool
schemas every turn, so prefill---quadratic in sequence length---dominates time-to-first-token
(TTFT) as the tool registry grows. Nexus's primary lever is to \emph{decouple routing from the
schema-prefill cost}: an INT8 semantic lookaside buffer (SLB) with a calibrated cross-encoder
margin gate selects tools by retrieval, and arguments are generated over a compressed textual
signature (median $19$ tokens) rather than over spliced key/value (KV) cache. This path is
depth-independent: routing accuracy stays near $89\%$ as the registry scales to $250$ tools---where
a concatenate-all-schemas baseline overflows the context window entirely---and it reaches a
first-argument token $1.66\times$ sooner than a full-schema re-prefill at an ${\approx}80\%$
main-context token saving. As a secondary, bounded lever we transplant a compiled schema KV block
directly into the live context. This is fundamentally limited by rotary position embedding (RoPE)
phase drift: an anchored splice is output-exact, but off-anchor placement corrupts attention, so
beyond a threshold $P=256$ Nexus repairs the seam with a depth-adaptive suffix redecode that
escalates to a full re-prefill. The resulting never-regress property is a guarantee on output
\emph{fidelity} (top-1 agreement, $D_{\mathrm{KL}}\approx 0$)---not on latency, which can dip to
$0.98\times$ before converging to parity---alongside a $1.1$--$1.7\times$ TTFT speedup at moderate
depth that narrows to parity at deep context. Two negative results bound the design: the off-anchor
RoPE fidelity boundary, and the failure of a reference-free drift gate to predict drift (Spearman
$\rho=0.193$). All measurements are from one model tuple (\code{Qwen2.5-14B-Instruct Q4\_K\_M}) on
Apple-silicon unified memory; the \emph{qualitative} boundaries generalize, while the
\emph{quantitative} envelope is tuple-specific.
\end{abstract}


%% ============================================================
\section{Introduction}
%% ============================================================
Tool-augmented LLM agents increasingly standardize on the Model Context
Protocol (MCP)~\cite{anthropic2024mcp}, presenting the model with a registry of callable
tools, each described by a verbose JSON schema. As the registry scales to hundreds of
tools, these schemas dominate the prompt. Because self-attention is $\mathcal{O}(N^2)$ in
sequence length $N$~\cite{vaswani2017attention}, re-encoding tool schemas every turn
imposes a prefill wall that scales poorly with the number and verbosity of available tools.

An appealing optimization is to compile each schema once into a KV block and transplant it
into the live context at inference time, paying the prefill cost only once. We find this is
fundamentally constrained by the position dependence of rotary position embeddings
(RoPE)~\cite{su2024roformer}: a block compiled for one absolute position cannot be relocated
to an arbitrary depth without corrupting attention. Prior serving systems sidestep the
problem by never moving cache (paged and prefix-shared KV~\cite{kwon2023vllm,zheng2024sglang}
keep blocks at the positions where they were computed). Nexus instead asks how far a
\emph{relocated} schema block can be trusted, and what it costs to repair it.

This paper presents the design, implementation, and measured envelope of \emph{Nexus}, a
user-space KV-cache serving prototype for agentic tool use on Apple-silicon unified memory
(UMA). Every mechanism below is anchored to a concrete symbol in the codebase, and every
number to a committed artifact under \code{results/v2.0\_canonical/}. We make four
contributions, tagged by evidence class:

\begin{enumerate}
  \item \textbf{Retrieval-decoupled routing} (\emph{measured}), the more durable and transferable
  result. Tool selection runs over an INT8 semantic lookaside buffer (SLB) with a calibrated
  cross-encoder gate, and arguments are generated over a compressed textual signature, avoiding the
  splice seam. It is depth-independent, scales to $250$ tools where the concatenate-all baseline
  cannot run, and delivers a $1.66\times$ first-argument latency win at ${\approx}80\%$ token saving
  (\S\ref{sec:arch}, \S\ref{sec:eval}).
  \item \textbf{Depth-adaptive recompute with a never-regress guarantee} (\emph{measured}), a
  bounded, characterized secondary lever. Past a $256$-token splice threshold, Nexus re-decodes a
  suffix fraction $R(n_{\mathrm{past}})$ that scales with depth to $100\%$, holding output fidelity
  exact and converging gracefully to prefill parity (\S\ref{sec:limits}, \S\ref{sec:eval}).
  \item \textbf{Systems mechanisms for UMA} (\emph{implemented}): transposed-V splicing for
  soft-capped attention models and cache-line hardening of the block allocator
  (\S\ref{sec:systems}).
  \item \textbf{Two bounding negative results} (\emph{well-supported}): the off-anchor
  RoPE fidelity boundary and the failure of reference-free drift gating (\S\ref{sec:limits}).
\end{enumerate}

We are explicit about scope. All benchmarks were executed on an Apple M4 Max SoC (16-core CPU, 40-core GPU, 16-core Neural Engine) equipped with 64\,GB Unified Memory and a 1\,TB NVMe SSD, running macOS/Darwin 25.5.0 (arm64), at git \code{1ce4aa4} / \code{llama.cpp cb2463bb}. All measurements use \code{Qwen2.5-14B-Instruct Q4\_K\_M} with \code{nomic-embed-text-v1.5}. Artifacts are under \code{results/v2.0\_canonical/}. The transposed-V splice mechanism is additionally implemented and fidelity-tested on a second, soft-capped model (\code{Gemma-2-9B}); that Gemma fidelity check lies outside the committed v2.0 artifact bundle. Physical splicing requires direct copies into local physical cache addresses, so it runs only for local, UMA-resident execution; cloud/remote APIs (which accept only text token streams) and discrete-GPU back-ends fall back entirely to text-prefill (Path~B, $1.0\times$ speedup). The precise architectural boundary is detailed in \S\ref{sec:vtrans}. Sample sizes are small ($n\!\le\!30$ for end-to-end arms), and the deep splice is validated at the fidelity-and-latency level, not as a multi-turn production path. We view the negative results and the measured envelope as the durable contribution.

Three research questions organize the paper. \textbf{RQ1}: how far can a relocated schema block be
trusted before its next-token distribution departs from a full prefill (\S\ref{sec:limits})?
\textbf{RQ2}: what does repairing that drift cost in TTFT (\S\ref{sec:eval})? \textbf{RQ3}: can
routing and argument generation avoid the splice path entirely (\S\ref{sec:arch},
\S\ref{sec:eval})?

\emph{Artifact availability.} Every reported number traces to a committed artifact under
\code{results/v2.0\_canonical/} at git \code{1ce4aa4} (\code{llama.cpp cb2463bb})---with the sole
exception of the placement-offset $D_{\mathrm{KL}}$ sweep of Fig.~\ref{fig:dkl}, regenerated at
\code{5d43008}---for the model
tuple \code{Qwen2.5-14B-Instruct Q4\_K\_M} with \code{nomic-embed-text-v1.5}. The artifact bundle
is available from the author on request pending release; a public archive with a DOI will be added
on publication.

%% ============================================================
\section{Related Work}
\label{sec:related}
%% ============================================================
\emph{Tool retrieval and schema compression.} The two standard responses to schema bloat both
act before the model. Retrieval-augmented tool selection---exemplified by
RAG-MCP~\cite{ragmcp2025}---fetches only the relevant MCP schemas prior to prefill. Schema-level
compilation~\cite{tscg2026} and learned latent tool retrieval~\cite{ntilc2026} shrink or replace
tool descriptions at the application and embedding levels, the latter requiring trained
alignment. Nexus adopts the same retrieval principle to decouple routing but adds a complementary
system-level lever: direct KV transplantation for the schemas that remain, with a zero-shot
calibrated margin gate over frozen embeddings. We position Nexus against RAG-MCP and in-place
KV-serving systems in Table~\ref{tab:comparison}; because those numbers are reported on different
hardware and workloads, we treat them as contextual rather than a controlled head-to-head, which
we leave to future work.

\emph{KV-cache management for serving.} Paged and prefix-shared KV
management~\cite{kwon2023vllm,zheng2024sglang} and dynamic virtual-memory
schemes~\cite{vattention2025,ellm} eliminate fragmentation and reuse cache \emph{in place}; they do
not relocate a block to a new position, which is precisely the regime Nexus characterizes.
RedKnot~\cite{redknot2026} pushes reuse to the granularity of individual attention heads but keeps
cache in place, whereas Nexus relocates coarse per-tool blocks. Fail-closed lowering of resident KV
claims~\cite{stepanek2026} formalizes when a runtime may treat a cached-KV obligation as satisfied;
Nexus's splicer takes the same fail-closed stance, declining transposed-V splices on discrete-GPU
back-ends rather than risk corruption. Prefill/decode disaggregation and chunked
prefill~\cite{agrawal2024sarathi,patel2024splitwise,zhong2024distserve} attack the same prefill cost
from the scheduling side and are complementary. Transactional speculative KV~\cite{transkvcache}
similarly separates stable from transient KV state.

\emph{RoPE, attention kernels, and decoding.} Context-window extension via frequency
interpolation~\cite{peng2023yarn} shares RoPE machinery with our reanchor path but targets longer
contexts, not cache transplantation. FlashAttention~\cite{dao2022flashattention,dao2023flashattention2}
determines the V-cache layout that dictates our transposed-V handling. Constrained
decoding~\cite{willard2023outlines,beurer2023lmql,guidance2023} underlies the FSM/JSON masking used
for argument generation. Nexus targets a different regime---relocating compiled schema blocks under
unified memory---and we make no outperformance claim against these systems. The SLB scan mechanism and \code{.atb} page alignment build directly on the neuro-symbolic memory substrate (Memory Palace/Atlas index, Trace DAG, SLB) introduced in Aeon v3~\cite{arslan2026aeon}, extending them to direct, zero-copy physical cache transplantation under UMA.

\begin{table*}[t]
\centering
\caption{Nexus in context. In-place KV-serving systems never relocate a compiled block; Nexus
relocates per-tool blocks and decouples routing from the splice path. Non-Nexus numbers are
reported on different hardware and workloads ($\dagger$: contextual, \emph{not} a controlled
head-to-head; see \code{results/comparison\_sources.md}).}
\label{tab:comparison}
\scriptsize
\begin{tabular}{@{}l l l p{4.5cm} p{3.3cm} c@{}}
\toprule
System & KV strategy & Routing & Reported metric (native HW/workload) & Workload / HW & H2H? \\
\midrule
RAG-MCP~\cite{ragmcp2025}                  & in-place (prompt-level)      & retrieve-$K$ + prefill & $43.1\%$ vs $13.6\%$ tool selection; ${>}50\%$ prompt tokens$^\dagger$ & MCP stress test, LLM eval & no \\
vLLM / PagedAttn~\cite{kwon2023vllm}       & in-place (paged)             & ---                    & $2\text{--}4\times$ throughput$^\dagger$                   & ShareGPT, A100/A10G       & no \\
SGLang / RadixAttn~\cite{zheng2024sglang}  & in-place (prefix reuse)      & ---                    & up to $6.4\times$ throughput$^\dagger$                     & LLM programs, A10G/A100   & no \\
vAttention~\cite{vattention2025}           & in-place (virtual contig.)   & ---                    & up to $1.23\times$ vs paged kernels$^\dagger$              & GPU serving               & no \\
RedKnot~\cite{redknot2026}                 & in-place (head-aware)        & ---                    & resource-efficiency gains (qualitative)$^\dagger$          & GPU serving               & no \\
\midrule
\textbf{Nexus} (this work)                 & \textbf{relocate} (per-tool) & SLB + margin gate      & routing $89\%$@250 tools; first-arg $1.66\times$; deep-splice $1.1\text{--}1.7\times$ & GitHub-MCP, M4 Max UMA, Qwen2.5-14B & --- \\
\bottomrule
\end{tabular}
\end{table*}

%% ============================================================
\section{Background and Motivation}
%% ============================================================
\subsection{Self-Attention and the Schema-Prefill Wall}
For a sequence of length $N$ and per-head dimension $d$, scaled dot-product attention
computes $\mathrm{softmax}(QK^\top/\sqrt{d})V$ with $Q,K,V\in\mathbb{R}^{N\times d}$,
incurring $\mathcal{O}(N^2 d)$ time and $\mathcal{O}(N^2)$ attention memory per layer
\cite{vaswani2017attention}. This expression is definitional, not a new bound; we cite it
only to locate the dominant cost. In agentic serving, $N=H+\sum_{k}S_k$ where $H$ is the
conversation history and $S_k$ the token length of the $k$-th tool schema. Because the
schemas are re-encoded every turn, reducing their prefill contribution is the dominant lever,
and the two standard baselines---concatenate all schemas, or retrieve $K$ and prefill
those---both pay this cost online.

\subsection{RoPE Phase Drift}
\label{sec:rope}
RoPE injects position by rotating query/key feature pairs by a position-dependent
angle~\cite{su2024roformer}. For feature pair $i$ with base frequency $\theta_i=b^{-2i/d}$
(the runtime model uses base $b=10^6$), a key compiled assuming anchor position $m_0$ but
consumed at position $n_{\mathrm{past}}$ accrues a phase error
\begin{equation}
\Delta\theta_i = (n_{\mathrm{past}}-m_0)\,\theta_i .
\label{eq:drift}
\end{equation}
This is a definitional consequence of RoPE, not an empirical predictor of the divergence
values in \S\ref{sec:limits}. Because attention scores depend on the \emph{relative} rotation
between query and key, a transplanted block whose keys carry the wrong absolute phase is read
off-axis by every subsequent query. The error grows with the placement offset
$(n_{\mathrm{past}}-m_0)$ and is non-uniform across dimensions (high-frequency pairs drift
fastest), so a pre-compiled schema block is faithful only in a neighborhood of its compile
anchor. The codebase makes this boundary concrete: the splicer computes
\code{delta\_pos = n\_past - header->base\_pos} (\code{nexus\_kv\_splicer.cpp}), and the
orchestrator declines a bare splice once $n_{\mathrm{past}}$ exceeds the splice cap
\code{max\_splice\_pos\_}${}=256$ (\code{nexus\_orchestrator.cpp}; surfaced as \code{MAX\_SPLICE\_POS}
in the Python agent).

Nexus counteracts the bulk of this error at splice time by \emph{reanchoring}: each
transplanted key is re-rotated from its compile anchor $m_0$ to the runtime position
$n_{\mathrm{past}}$---the exact inverse of the offset $\Delta\theta_i$ in
\eqref{eq:drift}---so that subsequent queries read the relocated block on-axis rather than
off it. Reanchoring recovers the \emph{relative} query--key rotation exactly, but the block's
keys and values were compiled under a different preceding context, so a small residual
divergence survives the correction. That residual---not the raw drift of \eqref{eq:drift}---is
what \S\ref{sec:limits} measures (the ${\sim}10^{-2}$-nat band of Fig.~\ref{fig:dkl}), and it
is what the depth-adaptive repair below drives to zero. This is why an off-anchor splice drifts
\emph{slightly} rather than catastrophically, and why the $256$-token boundary is a repair-start
rather than a hard failure point.

\subsection{Depth-Adaptive Recompute}
Rather than treat $256$ as a hard wall, Nexus repairs drift by re-decoding the trailing tokens
of the spliced schema, with a fraction that grows with depth. Let $M=256$ be the splice
threshold (\code{max\_splice\_pos\_}), $R_{\mathrm{base}}=5\%$ the baseline recompute fraction
(\code{recompute\_pct\_}), and $K$ the multiple of $M$ at which recompute reaches $100\%$
(\code{recompute\_full\_mult\_}). The effective fraction is
\begin{equation}
R(n_{\mathrm{past}})=
\begin{cases}
R_{\mathrm{base}}, & n_{\mathrm{past}}\le M,\\[2pt]
R_{\mathrm{base}}+\frac{n_{\mathrm{past}}-M}{M(K-1)}\bigl(100-R_{\mathrm{base}}\bigr), & n_{\mathrm{past}}>M,
\end{cases}
\label{eq:adaptive}
\end{equation}
clamped to $100\%$. This matches \code{eff\_pct()} in the code exactly. At deep positions the
fraction approaches $100\%$, which is numerically a full text prefill ($D_{\mathrm{KL}}(\mathbf{p}_0 \parallel \mathbf{p}_{\mathrm{splice}}) = 0$); $K$ is the
knob that trades TTFT flatness for how early full recompute engages.

%% ============================================================
\section{Physical Limitations and Negative Results}
\label{sec:limits}
%% ============================================================
We present the negative results first because they motivate the architecture: they fix the
shape of the repair curve and force routing to be depth-independent rather than to depend on a
deep splice.
\subsection{The Anchored/Off-Anchor Fidelity Boundary}
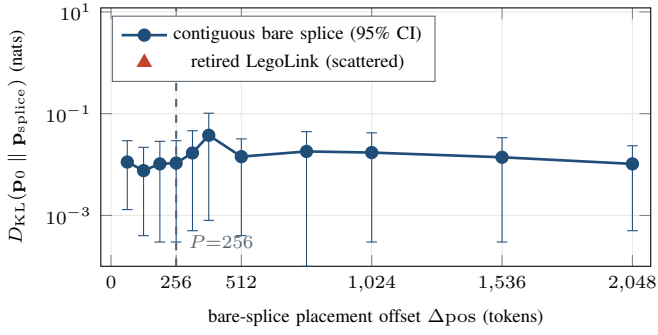
\begin{figure}[t]
\centering
\begin{tikzpicture}
\begin{axis}[
  width=\columnwidth, height=5.0cm,
  xlabel={bare-splice placement offset $\Delta\mathrm{pos}$ (tokens)},
  ylabel={$D_{\mathrm{KL}}(\mathbf{p}_0\parallel\mathbf{p}_{\mathrm{splice}})$ (nats)},
  ymode=log, ymin=1e-4, ymax=12,
  xmin=-40, xmax=2120, xtick={0,256,512,1024,1536,2048},
  legend style={font=\scriptsize,at={(0.02,0.97)},anchor=north west,draw=nxgray},
  grid=both, grid style={gray!18}, tick label style={font=\scriptsize}, label style={font=\scriptsize},
]
\addplot[nxblue,mark=*,line width=1pt,error bars/.cd,y dir=both,y explicit]
  table[row sep=\\,x=x,y=y,y error plus=ep,y error minus=em]{
  x y ep em \\
  64 0.0112 0.0182 0.0099 \\
  128 0.0076 0.0142 0.0072 \\
  192 0.0103 0.0183 0.0100 \\
  256 0.0107 0.0187 0.0104 \\
  320 0.0169 0.0292 0.0164 \\
  384 0.0376 0.0646 0.0368 \\
  512 0.0143 0.0176 0.0139 \\
  768 0.0181 0.0262 0.0180 \\
  1024 0.0172 0.0248 0.0169 \\
  1536 0.0139 0.0197 0.0136 \\
  2048 0.0103 0.0131 0.0098 \\
};
\addplot[nxred,mark=triangle*,only marks,mark size=3pt] coordinates {(1024,5.72)};
\draw[nxgray,dashed,line width=0.8pt] (axis cs:256,1e-4) -- (axis cs:256,12);
\node[nxgray,font=\scriptsize,anchor=south west] at (axis cs:270,1.5e-4) {$P{=}256$};
\legend{contiguous bare splice (95\% CI),retired LegoLink (scattered)}
\end{axis}
\end{tikzpicture}
\caption{Bare-splice next-token divergence vs.\ placement offset, recompute disabled
(\code{dkl\_sweep.json}; $10$ queries/offset, $95\%$ bootstrap CI). The contiguous-suffix splice
holds $D_{\mathrm{KL}}{\sim}10^{-2}$ nats with top-1 agreement $=1.0$ across the whole $0$--$2048$
range---two orders of magnitude below the retired LegoLink scattered-recompute point ($5.7$ nats at
$\Delta\mathrm{pos}{=}1024$). The enforced threshold $P{=}256$ (dashed) is a conservative
repair-start, not a fidelity cliff.}
\label{fig:dkl}
\end{figure}

We evaluate splice fidelity by the divergence of the model's next-token distribution from a
recompute-equivalent reference. At the compile anchor ($\Delta\mathrm{pos}=0$) an anchored
splice is output-exact: $D_{\mathrm{KL}}(\mathbf{p}_0 \parallel \mathbf{p}_{\mathrm{splice}}) \approx 0$ with $\text{top-1 agreement} = 1.0$. Once the
block is placed off-anchor, the phase error of \eqref{eq:drift} perturbs the distribution.
Fig.~\ref{fig:dkl} traces this perturbation directly: with recompute \emph{disabled} (a bare splice,
so the curve shows the raw off-anchor drift the repair must overcome), we sweep the placement offset
from the anchor to $\Delta\mathrm{pos}=2048$ and measure the next-token divergence against a
full-prefill reference at each depth ($10$ queries/offset, \code{dkl\_sweep.json}). Two facts stand
out. First, the RoPE reanchor is effective: the divergence rises from its ${\approx}0$ anchor floor to
${\sim}10^{-2}$ nats within the first $64$ tokens and then \emph{plateaus}---it stays in the
$0.008$--$0.038$-nat band, with top-1 agreement $=1.0$, across the entire $0$--$2048$ range, so a
contiguous-suffix bare splice never fails catastrophically at these depths. The threshold $P=256$ is
therefore a \emph{conservative} repair-start---where the drift has clearly separated from the anchor
floor while top-1 is still safe---rather than a sharp fidelity cliff.

Second, the retired scattered partial-recompute configuration (LegoLink) tells the complementary
cautionary story: at $n_{\mathrm{past}}=1024$ it left the divergence as high as
$D_{\mathrm{KL}}\approx 5.7$~nats (\code{rope\_boundary\_gate.json})---two orders of magnitude
\emph{above} the contiguous-suffix bare splice at the same depth (Fig.~\ref{fig:dkl}). Scattered
recompute thus actively corrupted the distribution rather than repairing it, which is precisely why
the current design repairs a \emph{contiguous} trailing suffix and why that mechanism is retired.
The production splicer still blocks bare splices beyond $n_{\mathrm{past}}=256$ and escalates the
repair fraction of \eqref{eq:adaptive} with depth. This raises a fair question: if the contiguous
bare splice already holds top-1 agreement $=1.0$ out to $\Delta\mathrm{pos}=2048$
(Fig.~\ref{fig:dkl}), why repair at all? Because top-1 agreement is a weaker property than
distributional identity. The residual ${\sim}10^{-2}$-nat divergence leaves the tail of the
next-token distribution perturbed, so under temperature or nucleus sampling the spliced and
prefilled paths can still diverge, and that per-token gap compounds across the many turns of an
agentic trajectory. Driving the divergence to \emph{exactly} $0$ (Table~\ref{tab:perf}) makes the
spliced output bit-identical to a full prefill---a guarantee that is auditable and holds under any
decoding parameters---which is what ``never-regress'' certifies. The depth-adaptive redecode buys
that hard guarantee for a little TTFT, rather than accepting a small-but-nonzero residual whose
downstream effect is workload-dependent.

\subsection{The Failure of Reference-Free Drift Gating}
\label{sec:gating}
A natural optimization for the Path-A ($\le 256$) to Path-B ($>256$) transition is a cheap,
reference-free runtime proxy---per-head preceding-context K-variance---that predicts when a deep
splice will drift catastrophically. Profiling this proxy against the true per-head drift on the
14B model (\code{gating\_nogo.json}) rules the approach out. The proxy does
not predict drift: mean Spearman $\rho=0.193$, well below the $0.40$ target. Per-head drift is not
itself absent---max drift ranges $175$--$207$ and mean drift $91$--$105$ across depths and both on-
and off-topic contexts (Table~\ref{tab:gating})---but the K-variance proxy fails to rank-correlate
with that variation, so a scalar-threshold gate driven by it cannot separate high- from low-drift
heads and would fire indiscriminately. An effective online gate would require a true
reference (the full prefill cost we seek to avoid) or a custom attention kernel. This justifies
keeping the deterministic depth-adaptive curve of \eqref{eq:adaptive} as the repair mechanism.

\begin{table}[t]
\centering
\caption{Reference-free drift gate: per-head drift varies (max $175$--$207$, mean $91$--$105$),
but the K-variance proxy fails to rank-correlate with it ($\rho=0.193$) (\code{gating\_nogo.json}).}
\label{tab:gating}
\small
\begin{tabular}{rlrrr}
\toprule
$n_{\mathrm{past}}$ & ctx & max drift & mean drift & Spearman $\rho$ \\
\midrule
256  & on  & 207.1 & 105.2 & 0.147 \\
256  & off & 195.3 &  97.8 & 0.221 \\
1024 & on  & 193.2 &  98.4 & 0.182 \\
1024 & off & 189.1 &  96.4 & 0.250 \\
2048 & on  & 194.7 &  97.6 & 0.159 \\
2048 & off & 175.4 &  91.0 & 0.197 \\
\midrule
\multicolumn{4}{l}{mean Spearman $\rho$ (target ${\ge}0.40$)} & \textbf{0.193} \\
\bottomrule
\end{tabular}
\end{table}

%% ============================================================
\section{Architecture: Retrieval-Decoupled Routing}
\label{sec:arch}
\begin{figure*}[t]
\centering
\begin{tikzpicture}[
  font=\small,
  node distance=10mm,
  box/.style={rounded corners=2pt,draw=nxgray,line width=0.5pt,align=center,
    inner sep=3pt,minimum height=8mm,fill=white},
  hot/.style={box,draw=nxblue,fill=nxblue!6},
  neg/.style={box,draw=nxred,fill=nxred!5},
  flow/.style={-{Stealth[length=2mm]},draw=nxgray,line width=0.6pt},
]
\node[box] (q) {User query};
\node[hot,right=of q] (emb) {Embed\\\scriptsize \code{nomic v1.5} (INT8)};
\node[hot,right=of emb] (slb) {SLB scan\\\scriptsize INT8 SIMD, top-$k$};
\node[hot,right=of slb] (gate) {Margin gate\\\scriptsize $\tau{=}0.0136$ (P20)};
\node[hot,right=of gate] (ce) {Cross-encoder\\\scriptsize rerank ($\sim$20\% fire)};
\node[hot,below=10mm of ce] (tool) {Resolved tool id};

\node[box,below=10mm of slb] (ir) {Semantic compressed IR\\\scriptsize signature, 19 tok (p50)};
\node[hot,left=of ir] (fsm) {FSM-masked\\arg generation\\\scriptsize JSON grammar};
\node[box,left=of fsm] (out) {Tool call\\\scriptsize name + args};

\node[hot,above=9mm of gate] (splice) {Anchored splice\\\scriptsize $n_{\mathrm{past}}{\le}256$, $D_{\mathrm{KL}}{\approx}0$};
\node[neg,above=9mm of ce] (deep) {Deep splice / re-prefill\\\scriptsize $n_{\mathrm{past}}{>}256$: $R(n_{\mathrm{past}})$};

\draw[flow] (q) -- (emb);
\draw[flow] (emb) -- (slb);
\draw[flow] (slb) -- (gate);
\draw[flow] (gate) -- (ce);
\draw[flow] (ce) -- (tool);
\draw[flow] (tool) -- (ir);
\draw[flow] (ir) -- (fsm);
\draw[flow] (fsm) -- (out);
\draw[flow,dashed,draw=nxblue] (gate) -- (splice);
\draw[flow,dashed,draw=nxred] (gate) -- (deep);
\begin{scope}[on background layer]
  \node[fit=(splice)(deep),draw=nxgray,dashed,rounded corners,inner sep=4pt,
    label={[nxgray,font=\scriptsize,anchor=south,yshift=1mm]above:KV-splice acceleration (schema prefill)}] {};
  \node[fit=(emb)(ce)(tool)(fsm),fill=nxteal!4,draw=nxteal!40,dashed,rounded corners,inner sep=6pt,
    label={[nxteal,font=\scriptsize,anchor=north,yshift=-1mm]below:Coarse re-entrant context lock (serialized turn execution for determinism)}] {};
\end{scope}
\end{tikzpicture}
\caption{Nexus request flow. Routing (blue, lower path) depends only on the query embedding and
the tool registry, never on the deep context, so it is depth-independent. KV-cache acceleration
(upper, dashed) is a separate concern: an anchored splice is output-exact for
$n_{\mathrm{past}}\le 256$, while beyond it the depth-adaptive repair $R(n_{\mathrm{past}})$ of
\eqref{eq:adaptive} escalates to a full re-prefill. Arguments are generated over a compressed
textual signature, not over spliced KV.}
\label{fig:arch}
\end{figure*}
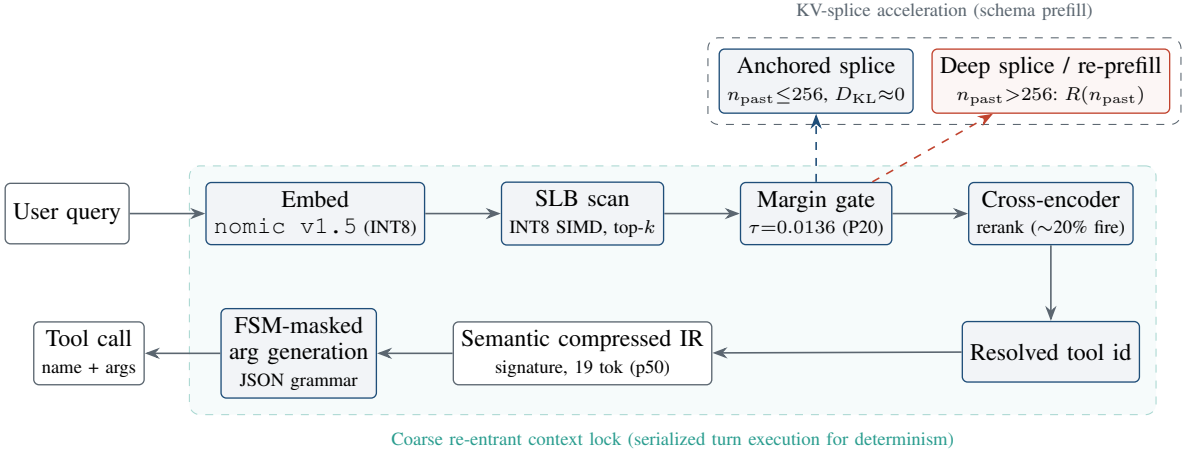

%% ============================================================
The negative results above dictate the architecture (Fig.~\ref{fig:arch}): routing must be
depth-independent rather than depend on a deep splice. Arguments, in turn, are generated over a
compact textual signature rather than over spliced schema KV, which both keeps the main-context
prompt small and avoids relying on a splice inside the drift-prone regime of \S\ref{sec:limits}.

\subsection{Semantic Lookaside Buffer and Gate}
Tool selection runs over a dense SLB (\code{nexus\_slb.cpp}): tool embeddings from
\code{nomic-embed-text-v1.5} are INT8-quantized and scanned with a branchless SIMD dot-product
(NEON / AVX-512-VNNI / AVX2). The embedding of a tool concatenates its name and description
under the asymmetric \code{search\_document:} prefix, mirrored by \code{search\_query:} on the
query side; the tool name carries dominant routing signal. The top candidates pass a margin gate:
if the top-two score margin exceeds a calibrated threshold the decision auto-routes, otherwise it
escalates to a fine-tuned MiniLM-class cross-encoder. The threshold is calibrated to the 20th
percentile of adversarial-pair margins, $\tau=0.0136$ (\code{calibration.json}), so the
cross-encoder fires on the lowest-confidence ${\sim}20\%$ of decisions. Retrieval depends only on
the query embedding and the registry, never on the deep context, so it is inherently
depth-independent.

\subsection{Execution Sidecar and Compressed IR}
When a candidate splice is required to validate routing in a fresh sequence, Nexus allocates a
temporary \emph{sidecar} sequence beginning at position $0$, so the anchored splice is always in
the output-exact regime of \S\ref{sec:limits}. To preserve multi-turn coreference (e.g., ``open a
pull request \emph{there}''), the sidecar carries a pruned sliding window: recent turns are kept
verbatim while bulky tool payloads are replaced by short semantic surrogates. Given the resolved
tool, arguments are generated in the main context over a \emph{semantic compressed intermediate
representation} (IR): the JSON
schema is reduced to a dense, type-hinted function signature with truncated inline descriptions,
e.g.\ \code{create\_repository(name: string, private?: boolean)}. At a median of $19$ tokens the IR
is far smaller than the full schema, so the main-context prefill avoids schema bloat while
retaining enough semantics to bind arguments to the correct fields---and, on the routed cases,
fills $100\%$ of query-specified arguments correctly (\S\ref{sec:eval}). A finite-state machine
masks logits to the tool-name radix trie (\code{nexus\_fsm.cpp}); JSON arguments are then constrained
by a native GBNF grammar compiled from the schema (\code{nexus\_agent.py}).

%% ============================================================
\section{Systems Implementation}
\label{sec:systems}
%% ============================================================
\subsection{Compiled Tool Blocks and Splice}
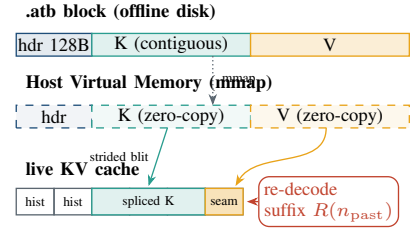
\begin{figure}[t]
\centering
\begin{tikzpicture}[font=\scriptsize,line width=0.5pt]
% .atb block on disk
\node[anchor=south west,align=left] at (0,2.45) {\textbf{.atb block (offline disk)}};
\draw[fill=nxblue!12,draw=nxblue] (0,2.1) rectangle (1.0,2.4) node[midway]{hdr 128B};
\draw[fill=nxteal!14,draw=nxteal] (1.0,2.1) rectangle (3.1,2.4) node[midway]{K (contiguous)};
\draw[fill=nxamber!18,draw=nxamber] (3.1,2.1) rectangle (5.2,2.4) node[midway]{V};

% Host Virtual Memory (mmap)
\node[anchor=south west,align=left] at (0,1.5) {\textbf{Host Virtual Memory (mmap)}};
\draw[fill=nxblue!6,draw=nxblue,dashed] (0,1.15) rectangle (1.0,1.45) node[midway]{hdr};
\draw[fill=nxteal!6,draw=nxteal,dashed] (1.0,1.15) rectangle (3.1,1.45) node[midway]{K (zero-copy)};
\draw[fill=nxamber!8,draw=nxamber,dashed] (3.1,1.15) rectangle (5.2,1.45) node[midway]{V (zero-copy)};

% live cache
\node[anchor=south west,align=left] at (0,0.4) {\textbf{live KV cache}};
\foreach \x in {0,...,5}{\draw[draw=nxgray] (\x*0.5,0.0) rectangle (\x*0.5+0.5,0.35);}
\node at (0.25,0.175){\tiny hist};\node at (0.75,0.175){\tiny hist};
\draw[fill=nxteal!14,draw=nxteal] (1.0,0.0) rectangle (2.5,0.35);
\node at (1.75,0.175){\tiny spliced K};
\draw[fill=nxamber!30,draw=nxamber] (2.5,0.0) rectangle (3.0,0.35);
\node at (2.75,0.175){\tiny seam};

% arrows
\draw[-{Stealth[length=1.5mm]},draw=nxgray,densely dotted] (2.6,2.1) -- (2.6,1.45) node[midway,right]{\tiny mmap};
\draw[-{Stealth[length=1.5mm]},draw=nxteal] (2.0,1.15) -- (1.75,0.38) node[midway,left]{\tiny strided blit};
\draw[-{Stealth[length=1.5mm]},draw=nxamber] (4.1,1.15) to[out=-90,in=60] (2.9,0.38);
\node[align=left,draw=nxred,rounded corners,inner sep=2pt,text=nxred,anchor=west] at (3.2,0.175)
  {re-decode\\suffix $R(n_{\mathrm{past}})$};
\draw[-{Stealth[length=1.5mm]},draw=nxred] (3.2,0.175) -- (3.05,0.175);
\end{tikzpicture}
\caption{Physical splice with memory mapping. The compiled block is mapped via zero-copy \code{mmap} into virtual memory. Contiguous F16 keys and values are then blitted into the live cache; the trailing suffix fraction $R(n_{\mathrm{past}})$ is re-decoded to stitch the seam. Layouts are transposed for soft-capped architectures (\S\ref{sec:vtrans}).}
\label{fig:atb}
\end{figure}

Each schema is compiled offline into an Aeon Tool Block (\code{.atb}): a $128$-byte header
recording the RoPE anchor and scaling parameters, followed by page-aligned ($2$\,MB) contiguous
F16 key and value tensors (\code{aeon\_tool\_block.hpp}). The header's RoPE parameters are validated
against the runtime model before any splice, so a block compiled for a different scaling regime is
rejected rather than silently corrupting attention. The splice writes K/V rows into the live KV
cells at the runtime cursor (Fig.~\ref{fig:atb}); a suffix of $\lceil S\cdot
R(n_{\mathrm{past}})/100\rceil$ tokens is then invalidated and re-decoded to stitch the seam
(\code{nexus\_orchestrator.cpp}).

\subsection{Transposed-V Splicing for Soft-Capped Attention}
\label{sec:vtrans}
Attention-logit soft-capping (e.g., \code{Gemma-2}) disables FlashAttention, which forces the V cache
into a transposed, token-innermost layout (\code{v\_trans=true}). A fixed-stride row-major copy
cannot address this layout. Nexus implements a layout-aware \code{splice\_v\_layer} routine
(\code{nexus\_kv\_splicer.cpp}) that performs strided transposed copies directly in the physical
cache under UMA, validates the live V strides before writing, and is fidelity-tested on the
soft-capped \code{Gemma-2-9B} model. On non-UMA / discrete-GPU
back-ends, where a transposed strided host-to-device transfer would be pathological, the splicer
declines and the system falls back to prefill---a deliberate safety boundary, not a general path.
The splice path therefore covers two GGUF regimes under local UMA: FlashAttention-compatible models
with a row-major V cache (e.g.\ \code{Qwen2.5}), spliced directly, and soft-capped models with a
transposed V cache (e.g.\ \code{Gemma-2-9B}), spliced via the layout-aware routine above. Two cases
fall back entirely to text-prefill (Path~B, $1.0\times$ speedup): cloud/remote execution, because
cache transplantation requires direct copies into local physical cache addresses that a token-stream
API does not expose; and discrete-GPU back-ends, where a transposed strided host-to-device transfer
would be pathological and the splicer declines by design.

\subsection{Cache-Line Hardening and Concurrency}
The Python orchestration layer drives a single native \code{llama\_context} whose internal decode
is guarded by a C++ mutex. Fine-grained Python locking around individual decode steps races that
mutex; Nexus instead serializes an entire turn (routing plus argument generation) under one coarse
re-entrant lock, yielding deterministic, bit-stable outputs under concurrency. To remove
cross-core false-sharing on the buddy allocator (\code{QuantizedBitmapAllocator}), the allocator is
declared \code{alignas(128)} (matching \code{NEXUS\_CACHE\_LINE}) and its contended mutex is pinned
to its own cache line, enforced at compile time:
\begin{center}\small
\code{static\_assert(alignof(QuantizedBitmapAllocator) \% 128 == 0);}
\end{center}
We report this as an implemented hardening with a passing four-thread determinism test; we do not
claim a throughput speedup, having measured none in isolation.

\subsection{L0 Exact-Token Radix Arena}
A zero-allocation radix cache (\code{nexus\_seq\_warm\_cache}) manages a fixed pool of $32$ sequence
slots and reuses warm prefixes by exact-token longest-common-prefix matching, copying KV within a
pre-reserved arena (\code{llama\_kv\_cache\_seq\_cp}) rather than allocating on the hot path.

%% ============================================================
\section{Evaluation}
\label{sec:eval}
%% ============================================================
All benchmarks were executed on an Apple M4 Max SoC (16-core CPU, 40-core GPU, 16-core Neural Engine) equipped with 64\,GB Unified Memory and a 1\,TB NVMe SSD, running macOS/Darwin 25.5.0 (arm64), at git \code{1ce4aa4} / \code{llama.cpp cb2463bb}. All measurements use \code{Qwen2.5-14B-Instruct Q4\_K\_M} with \code{nomic-embed-text-v1.5}. Artifacts are under \code{results/v2.0\_canonical/}. Sample sizes are stated per result; they are small, and we treat the numbers as an envelope rather than a population estimate. To make that framing concrete rather than a disclaimer, every headline proportion carries a Wilson $95\%$ interval and every TTFT median a bootstrap $95\%$ interval (the deep-splice cells use $n{=}15$ trials); speedup intervals propagate the prefill and splice median CIs. All quantitative splice
evaluation is restricted to the local, UMA-resident configuration of \S\ref{sec:vtrans};
back-ends outside that boundary fall back to text prefill with zero splice acceleration.

\subsection{Depth-Adaptive Splice: TTFT and Never-Regress}
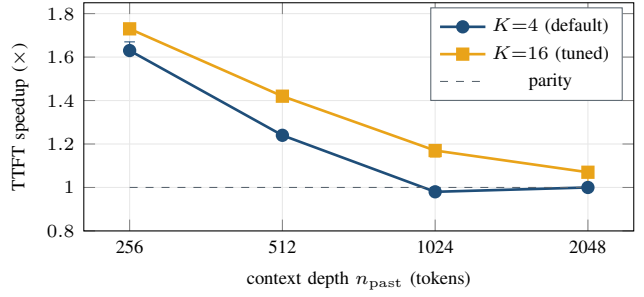
\begin{figure}[t]
\centering
\begin{tikzpicture}
\begin{axis}[
  width=\columnwidth, height=4.6cm,
  xlabel={context depth $n_{\mathrm{past}}$ (tokens)},
  ylabel={TTFT speedup ($\times$)},
  xmode=log, log basis x=2, xtick={256,512,1024,2048}, xticklabels={256,512,1024,2048},
  ymin=0.8, ymax=1.85, ytick={0.8,1.0,1.2,1.4,1.6,1.8},
  legend style={font=\scriptsize,at={(0.98,0.98)},anchor=north east,draw=nxgray},
  grid=both, grid style={gray!18}, tick label style={font=\scriptsize},
  label style={font=\scriptsize},
]
\addplot[color=nxblue,mark=*,line width=1pt,error bars/.cd,y dir=both,y explicit]
  table[row sep=\\,x=x,y=y,y error plus=ep,y error minus=em]{
  x y ep em \\ 256 1.63 0.04 0.01 \\ 512 1.24 0.01 0.01 \\ 1024 0.98 0.02 0.01 \\ 2048 1.00 0.005 0.01 \\};
\addplot[color=nxamber,mark=square*,line width=1pt,error bars/.cd,y dir=both,y explicit]
  table[row sep=\\,x=x,y=y,y error plus=ep,y error minus=em]{
  x y ep em \\ 256 1.73 0.02 0.01 \\ 512 1.42 0.01 0.005 \\ 1024 1.17 0.02 0.03 \\ 2048 1.07 0.01 0.06 \\};
\addplot[dashed,color=nxgray] coordinates {(256,1.0)(2048,1.0)};
\legend{$K{=}4$ (default),$K{=}16$ (tuned),parity}
\end{axis}
\end{tikzpicture}
\caption{TTFT speedup vs.\ depth for the two recompute curves (\code{deep\_splice\_ttft.json}).
Higher $K$ delays full recompute, sustaining speedup deeper at the cost of a larger repaired
suffix; both converge to parity as $R(n_{\mathrm{past}})\!\to\!100\%$.}
\label{fig:ttft}
\end{figure}

Table~\ref{tab:perf} reports median TTFT of the deep splice against a full text-prefill baseline,
at increasing context depth, for the default curve ($K=4$) and a flatter tuned curve ($K=16$),
with $15$ trials per cell (\code{deep\_splice\_ttft.json}). Two facts hold across every cell:
top-1 next-token agreement is preserved and the next-token $D_{\mathrm{KL}}(\mathbf{p}_0 \parallel \mathbf{p}_{\mathrm{splice}})$ is $\approx 0$---the never-regress guarantee.
We stress that never-regress is a guarantee on output \emph{fidelity} (top-1 agreement and
$D_{\mathrm{KL}}\approx 0$), not on latency: because the repaired suffix adds redecode work, measured
TTFT can dip slightly below parity (e.g.\ $0.98\times$ at $n_{\mathrm{past}}=1024$, $K=4$) before
converging to $1.0\times$; the fidelity guarantee is never violated even where the latency ratio is.
The speedup is not flat: at moderate depth the splice saves $1.1$--$1.7\times$, and as
$R(n_{\mathrm{past}})$ escalates toward $100\%$ the repaired splice converges to prefill parity
($\approx 1.0\times$). Flat TTFT and never-regress are in genuine tension, and $K$ is the dial
between them (Fig.~\ref{fig:ttft}).

\begin{table}[t]
\centering
\caption{Deep-splice TTFT vs.\ depth ($n{=}15$ trials/cell). Never-regress holds everywhere
(top-1 OK, $D_{\mathrm{KL}}(\mathbf{p}_0 \parallel \mathbf{p}_{\mathrm{splice}}) \approx 0$); speedup narrows to parity as recompute escalates.
Eq.~\eqref{eq:adaptive} reproduces the $R$(\%) column exactly (verified), so the recompute
schedule can be audited without the code. Speedup intervals propagate the per-cell bootstrap
$95\%$ CIs of the prefill and splice medians (independent timing loops). Source:
\code{deep\_splice\_ttft.json}.}
\label{tab:perf}
\footnotesize
\setlength{\tabcolsep}{3.5pt}
\begin{tabular}{crrrcc}
\toprule
$n_{\mathrm{past}}$ & $R$ (\%) & prefill (ms) & splice (ms) & speedup [95\% CI] & top-1/$D_{\mathrm{KL}}$ \\
\midrule
\multicolumn{6}{l}{\textbf{Default curve} $K=4$} \\
256  & 5.0   & 3278 & 2010 & $1.63$\,{\scriptsize[1.62,1.67]} & OK / $\approx 0$ \\
512  & 36.7  & 4672 & 3761 & $1.24$\,{\scriptsize[1.23,1.25]} & OK / $\approx 0$ \\
1024 & 100.0 & 7302 & 7425 & $0.98$\,{\scriptsize[0.97,1.00]} & OK / $0$ \\
2048 & 100.0 & 13092 & 13142 & $1.00$\,{\scriptsize[0.99,1.00]} & OK / $0$ \\
\midrule
\multicolumn{6}{l}{\textbf{Tuned curve} $K=16$} \\
256  & 5.0   & 3327 & 1924 & $1.73$\,{\scriptsize[1.72,1.75]} & OK / $\approx 0$ \\
512  & 11.3  & 4747 & 3336 & $1.42$\,{\scriptsize[1.42,1.43]} & OK / $\approx 0$ \\
1024 & 24.0  & 7625 & 6532 & $1.17$\,{\scriptsize[1.14,1.19]} & OK / $\approx 0$ \\
2048 & 49.3  & 13468 & 12620 & $1.07$\,{\scriptsize[1.01,1.08]} & OK / $\approx 0$ \\
\bottomrule
\end{tabular}
\end{table}

\subsection{Routing Accuracy and Registry Scale}
\begin{figure}[b]
\centering
\begin{tikzpicture}
\begin{axis}[
  width=\columnwidth, height=4.4cm,
  xlabel={registry size $N$ (tools)}, ylabel={accuracy (\%)},
  xmode=log, log basis x=10, xtick={10,50,100,250}, xticklabels={10,50,100,250},
  ymin=0, ymax=105, ytick={0,25,50,75,100},
  legend style={font=\scriptsize,at={(0.5,-0.28)},anchor=north,legend columns=2,draw=nxgray},
  grid=both, grid style={gray!18}, tick label style={font=\scriptsize}, label style={font=\scriptsize},
]
\addplot[color=nxblue,mark=*,line width=1pt,error bars/.cd,y dir=both,y explicit]
  table[row sep=\\,x=x,y=y,y error plus=ep,y error minus=em]{
  x y ep em \\ 10 92 3.9 7.0 \\ 50 90 4.5 7.4 \\ 100 89 4.7 7.6 \\ 250 89 4.7 7.6 \\};
\addplot[color=nxteal,mark=triangle*,line width=1pt,error bars/.cd,y dir=both,y explicit]
  table[row sep=\\,x=x,y=y,y error plus=ep,y error minus=em]{
  x y ep em \\ 10 76 7.3 9.2 \\ 50 74 7.6 9.4 \\ 100 74 7.6 9.4 \\ 250 74 7.6 9.4 \\};
\addplot[color=nxamber,mark=square*,densely dashed,line width=1pt] coordinates {(10,98)};
\addplot[color=nxred,mark=x,only marks,mark size=3pt] coordinates {(50,0)(100,0)(250,0)};
\legend{Nexus routing acc.,SLB R@1,oracle (fits),oracle: ctx overflow}
\end{axis}
\end{tikzpicture}
\caption{Routing accuracy vs.\ registry size (\code{routing\_accuracy\_n250.json}). Nexus stays
near $89\%$ as $N$ grows to $250$; the concatenate-all oracle is accurate at $N=10$ but overflows
the context window at $N\ge50$ and cannot route at all.}
\label{fig:route}
\end{figure}
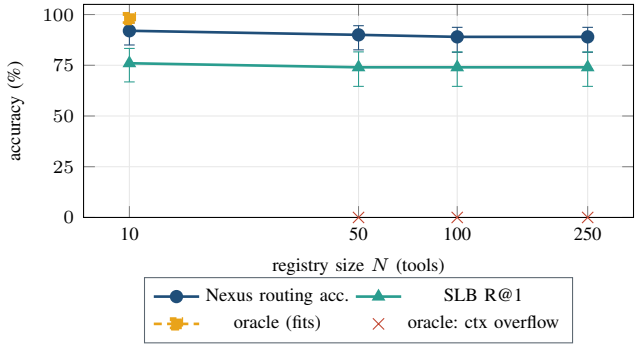

Routing is evaluated over $100$ GitHub-MCP queries as the registry scales from $N=10$ to $N=250$
tools (\code{routing\_accuracy\_n250.json}). End-to-end routing accuracy is nearly flat with scale:
$92\%$, $90\%$, $89\%$, $89\%$ at $N=10,50,100,250$ (Fig.~\ref{fig:route})---a spread that sits
within overlapping Wilson $95\%$ intervals, so ``nearly flat'' is a statement about the interval
band, not a claim that scale leaves accuracy strictly unchanged; at $N=250$ the routing
accuracy carries a Wilson $95\%$ CI of $[81.4, 93.7]$ ($n{=}100$), with SLB top-1 recall
$74\%$ $[64.6, 81.6]$ and top-3 recall $95\%$ $[88.8, 97.8]$. The contrast with the
concatenate-all-schemas oracle is
the point: the oracle reaches $98\%$ at $N=10$ but \emph{overflows the context window at $N\ge50$}
and cannot answer at all, whereas Nexus keeps the main-context prompt small and continues to route.
Because the oracle no longer fits the prompt at $N\ge50$, a token-reduction ratio against it is
undefined there; Nexus instead holds the main-context routing payload to a median $19$-token IR
(\S\ref{sec:arch}). Where a ratio is well defined---the sidecar's hybrid path against a
full-schema re-prefill---the main-context saving is ${\approx}80\%$ (\S\ref{sec:eval}). Because SLB search depends only on the query
embedding and the registry, tool count does not inflate TTFT: the in-situ INT8 scan over $250$ real
tool vectors completes in \SI{17.6}{\micro\second} (median, including the Python FFI boundary;
\code{routing\_accuracy\_n250.json}), while the pure C++ SIMD dot-product scan over $250$ unit-norm
vectors runs in \SI{8.25}{\micro\second} (median; \code{slb\_latency.json}). Registering the tool
name alongside its
description under the \code{nomic} prefixes is what makes retrieval scale: the tool name carries the
dominant routing signal, so omitting it degrades top-1 recall (a design rationale; no name-ablation
artifact is included in the v2.0 bundle). Table~\ref{tab:micro} summarizes routing and micro-costs.

\subsection{Argument Fidelity}
Over the compressed textual IR (median $19$ tokens, p99 $32$), the sidecar routes correctly on
$86.7\%$ (Wilson $95\%$ CI $[70.3, 94.7]$) of the $30$-case consistency set and, on the routed
cases, fills $100\%$ ($95\%$ CI ${\ge}91.2\%$ over $40$ specified arguments) of query-specified
arguments correctly with a $100\%$ JSON validity rate and no placeholder leakage
(\code{sidecar/accuracy.json}); end-to-end argument accuracy, which charges routing errors as
failures, is $80\%$ ($40/50$ arguments, $[67.0, 88.8]$). The hybrid path reaches its first-argument token in $443.8$~ms versus
$737.3$~ms for an oracle that re-prefills the full schema every turn---a $1.66\times$ reduction
at an $\approx 80\%$ main-context token saving. The pruned sliding window of \S\ref{sec:arch}
keeps recent-turn referents available for cross-turn arguments without re-admitting bulky schema
text; we do not report a separate coreference accuracy, as no such measurement is in the committed
v2.0 bundle.

\begin{table}[t]
\centering
\caption{Routing and micro-costs at $N{=}250$ tools. Proportions carry Wilson $95\%$ intervals
(bracketed); ${\ge}x$ denotes the lower bound of a $100\%$ cell. Sources:
\code{routing\_accuracy\_n250.json}, \code{slb\_latency.json}, \code{sidecar/accuracy.json},
\code{calibration.json}, \code{l0\_radix.json}.}
\label{tab:micro}
\footnotesize
\setlength{\tabcolsep}{3pt}
\begin{tabular}{lr}
\toprule
Quantity & Value \\
\midrule
Routing accuracy, $N{=}250$ ($n{=}100$)      & 89\,\% {\scriptsize[81.4,\,93.7]} \\
SLB top-1 / top-3 recall, $N{=}250$          & 74 / 95\,\% \\
SLB search latency, in-situ incl.\ FFI ($N{=}250$) & \SI{17.6}{\micro\second} \\
SLB search latency, pure C++ SIMD scan       & \SI{8.25}{\micro\second} \\
L0 radix copy latency (P50, $10$ seeds)      & \SI{3.04}{\micro\second} \\
L0 radix warm-hit rate ($10$ seeds)          & 69.5\,\% \\
Sidecar routing accuracy ($n{=}30$ cases)    & 86.7\,\% {\scriptsize[70.3,\,94.7]} \\
Specified-arg accuracy (routed, $n{=}40$ args) & 100\,\% {\scriptsize(${\ge}91.2$)} \\
End-to-end argument accuracy ($n{=}50$ args)  & 80\,\% {\scriptsize[67.0,\,88.8]} \\
JSON validity rate ($n{=}30$)                & 100\,\% {\scriptsize(${\ge}88.6$)} \\
IR length, median / p99                      & 19 / 32 tok \\
Cross-encoder gate $\tau$ (P20)              & 0.0136 \\
Gate fire rate (at $\tau{=}0.0136$)          & 20.8\,\% \\
Reference-free gate Spearman $\rho$          & 0.193 \\
\bottomrule
\end{tabular}
\end{table}

\subsection{Discussion}
The evaluation supports a bounded thesis. Off-anchor KV splicing is fundamentally limited by RoPE
phase drift, but the failure envelope is predictable and can be repaired by a deterministic
depth-adaptive recompute that never regresses below a prefill. On this workload that buys a
$1.1$--$1.7\times$ TTFT reduction at moderate depth, decaying to parity at deep context. Routing,
argument generation, and coreference are deliberately kept out of the splice path---a design forced
by the off-anchor fidelity boundary---and run over retrieval and a compact textual IR at
microsecond-scale overhead.

%% ============================================================
\section{Limitations and Scope}
\label{sec:scope}
%% ============================================================
We state the boundaries of these claims explicitly.
\emph{Single tuple.} All quantitative numbers are from one host (Apple M4 Max, UMA) and one
measured model tuple (\code{Qwen2.5-14B-Instruct Q4\_K\_M}) with one embedder and one \code{llama.cpp} build; the
transposed-V splice mechanism is additionally validated on \code{Gemma-2-9B}, but generality beyond
these is unproven.
\emph{Small $n$.} End-to-end arms use $n\le30$. Accuracy intervals are correspondingly wide and
we do not report them as population estimates.
\emph{Deep splice is micro-validated.} The never-regress curve is measured at the
fidelity-and-latency level (Table~\ref{tab:perf}); it is wired into the agent
(\code{enable\_deep\_splice}) but not exercised as a multi-turn production path.
\emph{Transposed-V is UMA-only.} Soft-capped models splice under unified memory; discrete-GPU
back-ends decline by design.
\emph{Retrieval ceiling.} Dense and cross-encoder recall have a workload-dependent ceiling; the
gate improves the lowest-confidence decisions but does not remove hard confusions.

%% ============================================================
\section{Conclusion}
%% ============================================================
Nexus reframes tool-schema acceleration from unconditional KV transplantation to
retrieval-decoupled routing paired with a depth-adaptive repair of the schema splice. We
characterized the RoPE-induced fidelity boundary in output-distribution terms, established the
off-anchor boundary and the reference-free gating failure as well-supported negative results.
We design the serving substrate to decouple semantic routing from the attention-level phase drift boundary,
executing partial recomputes only where mathematically required to restore output fidelity.
On a single-host 14B prototype this delivers a $1.1$--$1.7\times$ TTFT reduction at moderate depth
with high argument validity. We view the measured envelope and the negative results as the durable
contribution, and we separate what generalizes from what does not: the \emph{qualitative} boundaries
---off-anchor relocation drifts, scattered recompute cannot cheaply repair it, and reference-free
gating fails to predict it---are properties of RoPE and hold model-agnostically, whereas the
\emph{quantitative} envelope (the $P=256$ threshold, the recompute-curve constants, and the specific
divergence magnitudes) is calibrated to one model tuple and must be re-measured elsewhere.

\bibliographystyle{IEEEtran}
\bibliography{references}

\end{document}